\documentclass[journal]{IEEEtran}

\usepackage{soul,color}

\usepackage{cite}

\ifCLASSINFOpdf
   \usepackage[pdftex]{graphicx}
  \graphicspath{{Figures/}}
  \DeclareGraphicsExtensions{.pdf,.jpeg,.png}
\else
\fi
\usepackage{fancyhdr}
\begin{document}
\onecolumn
\thispagestyle{empty}\setcounter{page}{0}%
\vfill
{\Large	 \textcopyright 2021 IEEE.  Personal use of this material is permitted.  Permission from IEEE must be obtained for all other uses, in any current or future media, including reprinting/republishing this material for advertising or promotional purposes, creating new collective works, for resale or redistribution to servers or lists, or reuse of any copyrighted component of this work in other works.
\\ \\
This article has been accepted for publication in IEEE Transactions on Medical Robotics and Bionics.\\ \\
\textbf{DOI:} 10.1109/TMRB.2021.3073209\\ \\
\textbf{URL:} https://ieeexplore.ieee.org/document/9404322}
\twocolumn

%
\title{Rate of Orientation Change as a New Metric for Robot-Assisted and Open Surgical Skill Evaluation}
%
%

\author{Yarden~Sharon,~\IEEEmembership{Student Member,~IEEE,}
        Anthony~M.~Jarc,
        Thomas~S.~Lendvay,
        and~Ilana~Nisky,~\IEEEmembership{Senior Member,~IEEE}
        
\thanks{Manuscript revised \_\_\_}        
\thanks{This research was supported by the Helmsley Charitable Trust through the ABC Robotics Initiative and by the Marcus Endowment Fund both at Ben-Gurion University of the Negev, the ISF grant number 327/20, the Israeli Ministry of Science and Technology grant 15627-3 and a grant for the Israel Italy Virtual Lab on artificial somatosensation for humans and humanoids. Y. Sharon was supported by the Besor scholarship and the Israeli Planning and Budgeting Committee scholarship.}
\thanks{Y. Sharon and I. Nisky are with the Department of Biomedical Engineering and Zlotowski Center for Neuroscience, Ben-Gurion University of the Negev, Beer-Sheva, Israel (e-mail: shayar@post.bgu.ac.il and nisky@bgu.ac.il).}
\thanks{A. Jarc is with Medical Research, Intuitive Surgical Inc., Norcross, GA, USA (e-mail: anthony.Jarc@intusurg.com).}
\thanks{T. S. Lendvay is with the Department of Urology, University of Washington, Seattle, WA, USA (e-mail: thomas.lendvay@seattlechildrens.org).}}

%
%

\markboth{}
{}
%



\maketitle
\thispagestyle{fancy}

\begin{abstract}
Surgeons’ technical skill directly impacts patient outcomes. To date, the angular motion of the instruments has been largely overlooked in objective skill evaluation. To fill this gap, we have developed metrics for surgical skill evaluation that are based on the orientation of surgical instruments. We tested our new metrics on two datasets with different conditions: (1) a dataset of experienced robotic surgeons and nonmedical users performing needle-driving on a dry lab model, and (2) a small dataset of suturing movements performed by surgeons training on a porcine model. We evaluated the performance of our new metrics (angular displacement and the rate of orientation change) alongside the performances of classical metrics (task time and path length). We calculated each metric on different segments of the movement. Our results highlighted the importance of segmentation rather than calculating the metrics on the entire movement. Our new metric, the rate of orientation change, showed statistically significant differences between experienced surgeons and nonmedical users / novice surgeons, which were consistent with the classical task time metric. The rate of orientation change captures technical aspects that are taught during surgeons' training, and together with classical metrics can lead to a more comprehensive discrimination of skills. \end{abstract}


%
\IEEEpeerreviewmaketitle

\section{Introduction}
%
%
%
%
\IEEEPARstart{S}{uccessful} surgery requires cognitive skill -- "knowing what to do", and motor skill -- "knowing how to do it" \cite{ericssonDeliberatePracticeAcquisition2004}. The technical skill of a surgeon directly impacts patient outcomes \cite{birkmeyerSurgicalSkillComplication2013}. Training programs are intended to bring junior surgeons to a high level of procedural and technical skill, but because of limited standard technical skill metrics, the maintenance of certification for practicing surgeons is mostly cognitive-based. For both cognitive and motor goals, it is paramount to evaluate the quality of surgeon's performance.

State of the art surgical skill assessment is still largely based on direct or video observation by expert surgeons. However, such evaluation is problematic for several reasons. First, subjective assessment may vary between evaluators \cite{reznickTeachingTestingTechnical1993}, and suffer from bias. Second, even if the assessment is structured using checklists \cite{martinObjectiveStructuredAssessment1997,vassiliouGlobalAssessmentTool2005,gohGlobalEvaluativeAssessment2012}, it is still limited by what the observers see and by their attention. Third, these observations require time. Therefore, it is important to find objective metrics that can describe the surgical performance in detail. Such metrics can help to identify training deficiencies more accurately, and to provide the trainee with valuable near real-time feedback to optimize their performance \cite{enaniWhatAreTraining2017}. 

The simplest objective metric is task completion time \cite{vanrijCusumAidEarly1995}. However, task time does not provide information about the quality of the action \cite{moorthyObjectiveAssessmentTechnical2003}. For example, a task that was completed fast might have been accomplished with careless instrument gestures. Development of more complex objective metrics is now possible, thanks to the advancement of technology. Tracking systems \cite{dattaUseElectromagneticMotion2001,trejosSensorizedInstrumentSkills2009} and virtual reality trainers \cite{wilsonMISTVRVirtual1997,eriksenObjectiveAssessmentLaparoscopic2005} enable the collection of motion information and the calculation of objective metrics such as path length of the instruments, number of movements, speed of movements, and number of errors \cite{thijssenContemporaryVirtualReality2010,reileyReviewMethodsObjective2011,vedulaObjectiveAssessmentSurgical2017,oquendoAutomaticallyRatingTrainee2018,hungDevelopmentValidationObjective2018}. However, to reach accurate surgical skill assessment, there is still much room for improvement.

One of these enabling technologies is teleoperated robot-assisted minimally-invasive surgery (RMIS). In RMIS, the surgeon teleoperates robotic surgical instruments inside the body of the patient \cite{maesoEfficacyVinciSurgical2010}. This allows for both unobtrusive tracking of the movements of the surgeon, e.g. position, velocity, and angular velocity, and using this information to evaluate skill \cite{vemerMeasurementsLevelSurgical2003,hernandezQualitativeQuantitativeAnalysis2004,niskyUncontrolledManifoldAnalysis2014,estradaSmoothnessSurgicalTool2016,frenchPredictingSurgicalSkill2017}. This abundance of data also motivates the use of advanced machine learning techniques for skill evaluation \cite{kumarAssessingSystemOperation2012,wangDeepLearningConvolutional2018}. However, such classification techniques are limited in terms of their ability to provide the trainee with meaningful feedback. 

To date, the orientation of surgical instruments has been used for calculate the angular path around different axes of the laparoscopic tool for assessing laparoscopic skill  \cite{eriksenObjectiveAssessmentLaparoscopic2005,hogleDocumentingLearningCurve2007,kundhalPsychomotorPerformanceMeasured2009,hofstadStudyPsychomotorSkills2013}. However, for robotic and open techniques, where the movement is not limited to rotations around certain axes, there has been no extensive use of orientation-based metrics for skill evaluation. This is somewhat surprising, because rotation of instruments is critical in many surgical tasks. For example, in needle-driving, surgeons are taught to rotate their wrist so that the needle addresses the tissue at a right angle and pierces the tissue with the least amount of force. Previous studies found that the angular velocity of the hands of experts was significantly higher compared to novices \cite{vemerMeasurementsLevelSurgical2003,oleynikovEffectVisualFeedback2006}, but this measurement was not linked to a specific task. In sports, measures of rotation have been used to assess skill among tennis players \cite{ahmadiWearableDeviceSkill2009}. Therefore, we believe that orientation-based metrics can also be used for evaluating surgical skill. 

In this study, we developed orientation-based metrics for surgical skill assessment in robot-assisted and open needle-driving movements. These metrics capture critical aspects of surgical expertise that, to the best of our knowledge, were not quantified by the existing metrics for surgical skill. Our metrics were designed to assess movements in which the DOF of the hand are not limited by the surgical instrument; therefore, we did not include conventional laparoscopic movements. We chose the needle-driving task as a good example, characterized by a combination of high clinical importance, technical complexity, and minimal necessary procedural knowledge. Moreover, needle-driving is the building block of surgical suturing that is part of the majority of surgical procedures regardless of the specialty field \cite{aggarwalTrainingJuniorOperative2006}. 

To investigate the performance of our new orientation-based metrics, we used two datasets. The first dataset, named here "Dry Lab", was collected in a previous study that compared teleoperated and open unimanual needle-driving movements of experienced robotic surgeons and nonmedical users \cite{niskyTeleoperatedOpenNeedle2015}. The teleoperated needle-driving was performed using the da Vinci Research Kit (dVRK) \cite{kazanzidesfOpensourceResearchKit2014}, which is a custom research version of the da Vinci Surgical System. The open needle-driving was performed with a needle-driver that was equipped  with magnetic trackers. Because each stage of the task has different constraints, different metrics may be required for the different stages. Therefore, we hypothesized that segmentation of the needle-driving movement into its stages is necessary prior to calculation of metrics in order to assess surgical skills. Specifically, we assumed that during the part of the needle insertion when rotation motion is required, orientation-based metrics can highlight differences between different levels of surgical skill.

The second dataset, named here "Porcine", was collected during the training of surgeons on a porcine model, and consisted of a series of tasks that highlight technical skills when using the da Vinci Surgical System, such as knot tying and third arm retraction. The analysis of this dataset is used to demonstrate the feasibility of using our orientation-based metrics in the analysis of realistic surgical data.

A preliminary version of this study for a subset of metrics on the teleoperated movements of the Dry Lab dataset was presented in an extended abstract form \cite{sharonInstrumentTipAngular2016}. 

\section{Methods}
The novelty of this study is the development and examination of orientation-based metrics for surgical skill evaluation. We were interested in the development of robust metrics whose performance is not dependent on experimental conditions. Therefore, we chose to test our metrics using data from several experiments with different conditions. In this section, we first briefly present the experimental setup, data acquisition, preprocessing, and segmentation of each dataset. Then, we present the calculation of the metrics, and the statistical analysis. We use $\mathbf{x}$ as the Cartesian translation vector ($x,y,z$ position coordinates), and $\varphi$ as the opening angle of the needle-driver. To present orientations, we use $\mathbf{R}$ for the rotation matrix that consists of three unit vectors ($\mathbf{R}=[\mathbf{\hat{x}},\mathbf{\hat{y}},\mathbf{\hat{z}]}$), and $\mathbf{Q}$ for the quaternion that consists of four components ($\mathbf{Q}=[q_1,q_2,q_3,q_4]$). The $^P$ superscript stands for PSM, and $^O$ for open needle-driver. $j$ is the index of sampled data points. 

\subsection{Dataset 1: Dry Lab}
\subsubsection{da Vinci Research Kit Setup}
Full details of the experimental setup and procedures are reported in \cite{niskyTeleoperatedOpenNeedle2015}, but we summarize here the important details for the current study. The setup of the dVRK that was used in the experiment is depicted in Fig. \ref{fig_setup}(a-b). The system consisted of a pair of Master Tool Manipulators (MTMs), a pair of Patient Side Manipulators (PSMs), a high resolution stereo viewer, and a foot-pedal tray. Two large needle-drivers were used as PSM instruments. Using the stereo viewer, the participant watched a 3D view of the task scene. A pair of Flea 3 (Point Grey, Richmond, BC) cameras were mounted on a custom designed fixture. The position and orientation of the camera were manually adjusted to obtain the best view of the task board, and were fixed throughout all the experiments.

The teleoperation was implemented as a position-exchange with PD controllers. The Cartesian position and the orientation of the tooltips were calculated from the sampled joint angles via forward kinematics. Velocities were calculated using numerical differentiation and filtering with a second-order Butterworth low-pass filter with a 20 Hz cutoff. To control the PSM, the position and velocity of the MTM were down-scaled by factor of 3 to mimic the `fine' movement scaling mode of the clinical da Vinci system, the orientation was not scaled. Similarly to the clinical da Vinci, there was no force feedback, and there was a small torque feedback on the orientation degrees of freedom to help users avoid large misalignment in tool orientation between the PSM and MTM.

\subsubsection{Experimental Procedures}
Sixteen participants performed the experiment that was approved by the Stanford University Institutional Review Board, after giving informed consent. The participants included six experienced surgeons (five urologists, $n\textsubscript{robotic procedures}>120$, and one general surgeon, $n\textsubscript{robotic procedures}>150$, self-reported), and ten nonmedical users (engineering graduate students) without surgical experience. There were no restrictions on the handedness of the surgeons, and the nonmedical users were all right handed by self-report. One nonmedical participant had extensive experience with the experimental setup, and hence was removed from the analysis.

\begin{figure}[!t]
\centering
\includegraphics[width=3.5in]{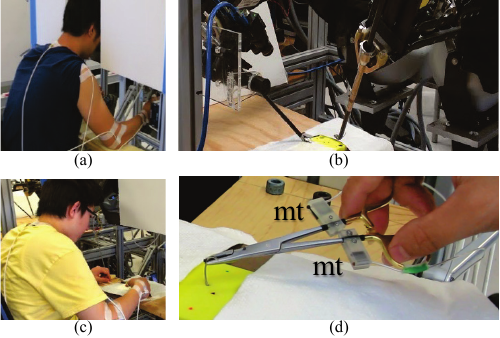}
\caption{Teleoperated and open needle-driving with the da Vinci Research Kit (dVRK). (a) A participant in the teleoperated session, seated at the master console. (b) The task board and the instruments on the patient-side table. (c)  A participant in the open session seated in front of the task board. (d) A surgical needle-driver with magnetic trackers ("mt").}
\label{fig_setup}
\end{figure}

\begin{figure}[!t]
\centering
\includegraphics[width=3.5in]{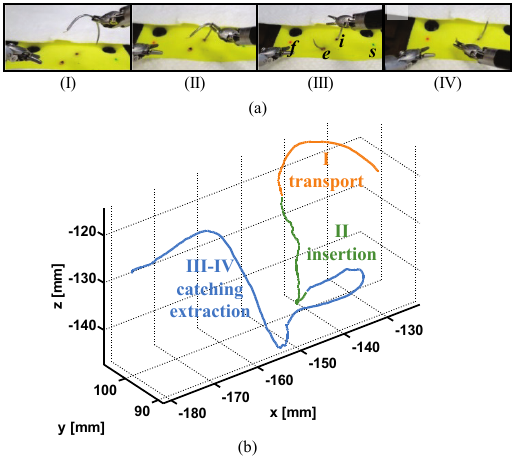}
\caption{Task segments. (a) The task board and task segments: (I) transport, (II) insertion, (III) catching, (IV) extraction. (b) An example of the path of an experienced surgeon. The numbers and the different colors indicate the segments of the task.}
\label{fig_task_segments}
\end{figure}

Each participant performed both teleoperated and open needle-driving sessions. The order of the two sessions was balanced across participants, i.e., half of the experienced surgeons and half of the nonmedical users performed the robotic task first, the other halves performed the open task first. The assignment into these groups was random. The participants performed the teleoperated needle-driving using the dVRK with a large needle-driver (Fig. \ref{fig_setup}(a-b)), and the open needle-driving using standard surgical needle-driver (Fig. \ref{fig_setup}(c-d)). The needle was a CT-1 tapered needle without its thread. Each task board consisted of four identical sets of targets, but only one set of targets was visible during a particular trial. Each set of targets on the task board consisted of four marks (Fig. \ref{fig_task_segments}(a).III): start (\textit{s}), insertion (\textit{i}), exit (\textit{e}), and finish (\textit{f}). 

In the teleoperated session, the participant sat in front of the master console of the dVRK (Fig. \ref{fig_setup}(a)). The task board was rigidly mounted on the patient-side table (Fig. \ref{fig_setup}(b)), such that its position was fixed relative to the cameras. 
In the open session (Fig. \ref{fig_setup}(c)), to provide a similar context to the teleoperated session, the participant also sat in front of the dVRK. A similar task-board was mounted on the armrest of the dVRK. To determine the position and orientation of the surgical needle-driver's tip, magnetic pose trackers (trakSTAR, Ascension Technology Corporation, Shelburne, VT) were mounted on its shafts (Fig. \ref{fig_setup}(d)). To prevent signal distortion, the tracker was separated from the metal body of the driver by 2 cm.

Each participant watched an instructions video before each session (teleoperated or open). Each session included 80 trials; after each block (10 trials) a break was offered. After two blocks, the suture-pad was readjusted so that a fresh area of the pad and targets were presented to the participant. Each trial started with a bimanual  adjustment of the needle in the right needle-driver in a configuration that is appropriate for the task. Then, participants placed the tip of the needle at start target (\textit{s}), and pressed the left foot-paddle (teleoperated) or mouse-button (open) to indicate the beginning of the task sequence. 
A single needle-driving trial consisted of four stages, as depicted in the video and in Fig. \ref{fig_task_segments}(a): (I) transport -- reaching with needle head from \textit{s} to \textit{i}, (II) insertion -- driving the needle through the tissue until its tip exits at \textit{e}, (III) catching -- opening the needle-driver and catching the tip of the needle, and (IV) extraction -- pulling the needle and reaching to \textit{f} with its tail. The trial ended when the tail of the needle was placed at the end target, and the left foot-paddle or mouse-button were pressed to indicate trial end.
During the experiment, some of the trials were not performed according to the instructions or not recorded properly. These trials were identified during the experiment prior to data analysis, and were removed from the analysis. Among teleoperated sessions, 28 out of 1200 trials were removed, and in the open sessions, 58 trials were removed. 

\subsubsection{Data Acquisition and Preprocessing}\label{dVRKOpenDataAcquisition}
In the teleoperated session, we analyzed the right PSM's data. The Cartesian position, velocity, orientation and opening angle of the needle-driver were recorded at 2 kHz. In the open session, the position and the orientation of the two magnetic pose trackers were recorded at 120 Hz. 

\begin{figure}[!t]
\centering
\includegraphics[width=3.5in]{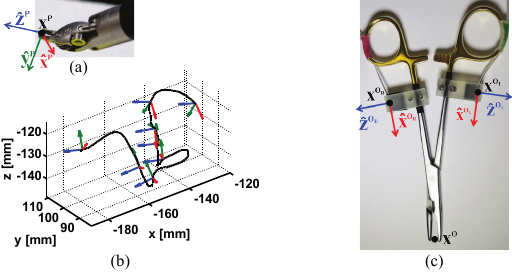}
\caption{Orientation reference frames. (a) Orientation reference frame of the PSM. (b) Right PSM's orientations at several samples along the path. (c) Orientation reference frames of the right and left magnetic trackers on the needle-driver.}
\label{fig_orientation}
\end{figure}

We interpolated and downsampled all data to 100 Hz. We filtered the Cartesian position offline at 6 Hz with a 2\textsuperscript{nd} order zero lag low-pass Butterworth filter using the Matlab function \texttt{filtfilt()}. In the open condition, we calculated the mapping from the position of the sensors $\mathbf{x}^{O_R}$ and $\mathbf{x}^{O_L}$ to the driver's endpoint  $\mathbf{x}^{O}$ (Fig. \ref{fig_orientation}(c)) using a calibration dataset. We calculated the Cartesian velocity using numerical differentiation of the filtered position.

In both conditions, the orientations were recorded as rotation matrices. Rotation matrices are orthogonal by definition, but because the resolution of the recording is limited, the recorded matrices were not orthogonal. Therefore, we used singular value decomposition (SVD) to find the nearest orthogonal matrix for each sampling point. Then, we converted the matrices to quaternions using the Matlab \texttt{dcm2quat()} function, and interpolated them using spherical linear interpolation (SLERP) \cite{corkeRoboticsVisionControl2011}. Quaternions that represent orientation are unit quaternions (i.e., normalized quaternions), and therefore, we normalized the quaternions after each calculation. 

In the open condition, the opening angle of the driver, $\varphi^O$, was calculated as $\varphi^O=cos^{-1}(\mathbf{\hat{x}}^{O_R}\cdot\mathbf{\hat{x}}^{O_L})$, where $\mathbf{\hat{x}}^{O_R}$ and $\mathbf{\hat{x}}^{O_L}$ are elements from the rotation matrices which represent the orientations of the right and left magnetic trackers (Fig. \ref{fig_orientation}(c)), and  $\cdot$ is the dot product. In the teleoperated condition, the opening angle
was available directly from the recorded data.

\subsubsection{Segmentation}
We segmented the needle-driving movement into four stages (Fig. \ref{fig_task_segments}(b)). Since there was no video recording of the movements, we segmented the movements using the kinematic trajectories. Because the movements were structured, we could define the transition between segments by specific indicators in the recorded signals, such as a minimum of the velocity or a threshold of the opening angle. To replace the manual search for these indicators, we built an algorithm that automatically finds them. The segmentation of all the trials was then checked visually, and when the algorithm failed to segment the movement, we corrected the segmentation manually. This happened in 7 teleoperated trials, and in 127 open trials, including all the trials of one of the participants.

The movement's trajectory and the opening angle were helpful for the segmentation of the first two segments, until the participant opened the needle-driver for the first time. However, after these two segments, the movements were very different from each other, so we could not define the transition between the segments using the kinematics signals. For instance, using the kinematics of the tools we could not identify a situation in which the participant was not able to pull the needle out, and tried to insert it again. Therefore, in this paper, we focus on the first and second segments, which were relatively consistent across participants and trials.

\subsection{Dataset 2: Porcine}
\subsubsection{Setup} Data were collected from da Vinci Xi surgical systems (Intuitive Surgical, Inc.) using an Intuitive Data Recorder (IDR). The data consisted of a single channel of endoscopic video and kinematics based on the joint angles of the patient side cart.

\subsubsection{Experimental Procedures}
\begin{figure}[!t]
\centering
\includegraphics[width=3.5in]{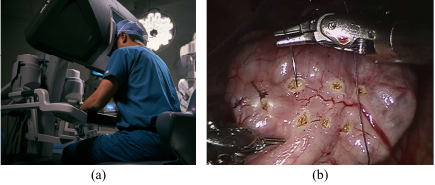}
\caption{The Porcine dataset. (a) A surgeon at the da Vinci surgical system surgeon console, image from: www.intuitive.com. (b) The suturing task on a porcine model.}
\label{fig_ClinicalLike}
\end{figure}

Three experienced ($n\textsubscript{robotic procedures}>200$) and four novice ($n\textsubscript{robotic procedures}<100$) surgeons completed a clinical-like suturing task on a porcine model that targeted the technical skills of using the da Vinci system (Fig. \ref{fig_ClinicalLike}). The suturing exercise required surgeons to use a two-hand technique to tie 4 interrupted sutures using large needle-drivers. The suturing task was part of a series of clinical-like activities conducted by each surgeon.

\subsubsection{Data Acquisition and Preprocessing}
In this dataset, we used the videos from the endoscopic camera for segmentation, and the PSM's data for calculation of the metrics. The videos were recorded at 30 frames per second. The participants were not limited to performing the task with one specific hand, and therefore, we analyzed the data of the tool that was used in each specific segment. The Cartesian position and orientation of the tools were recorded at 50 Hz. The filtering of the Cartesian position data, and the conversion of rotation matrices to quaternions was performed similarly to the analysis of the Dry Lab dataset, as described in \ref{dVRKOpenDataAcquisition}.

\subsubsection{Segmentation} The part of the movement that was consistent across participants and attempts was when the surgeon inserts the needle into the tissue. Therefore, we chose to analyze only the data of the insertion part to enable comparisons between the movements. We used the video stream to manually label these segments. Each participant performed four or five sutures, for every suture we isolated the insertion segment. When there were two attempts of needle insertion for the same suture, we included both attempts in the data analysis, and it happened only twice in the entire dataset.

\subsection{Metrics}
For each trial and each segment, we calculated four metrics: (1) task time -- the time elapsed between the beginning and the end of the movement; (2) path length -- the distance travelled by the instrument; (3) angular displacement -- the accumulated change in instrument orientation; and (4) rate of orientation change -- the average rate of the change in instrument orientation.
The first two are classical metrics for skill evaluation. We included these metrics to allow us to compare the performance of our new orientation-based metrics to the classical ones.

\begin{figure}[!t]
\centering
\includegraphics[width=3.5in]{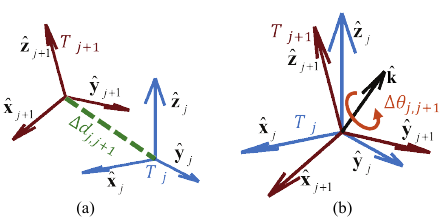}
\caption{Transformation between consecutive sampled frames $T_j$ and $T_{j+1}$. (a) Distance$\Delta d_{j,j+1}$. (b) Rotation around the axis $\mathbf{\hat{k}}$ by an angle $\Delta \theta_{j,j+1}$}
\label{fig_transformation}
\end{figure}

The \textit{task time} was calculated as:
\begin{equation}
    TT=t_{1_{i+1}}-t_{1_{i}},
\end{equation}
where $t_{1_{i}}$ is the time elapsed between the beginning of the movement and the beginning of the i\textsuperscript{th} segment. Then, we found the distance $\Delta d_{j,j+1}$ between pairs of consecutive sampled frames at the instrument's tip, $T_j$ and $T_{j+1}$ (Fig. \ref{fig_transformation}(a)):
\begin{equation}
    \Delta d_{j,j+1}=||\Delta x_j,\Delta y_j,\Delta z_j||,
\end{equation}
where $\Delta x,\Delta y,\Delta z$ are the differences in the $x,y,z$ positions, respectively. Using the distance, we calculated the \textit{path length} for the N samples of the segment:
\begin{equation}
    P=\sum_{j=1}^{N-1} \Delta d_{j,j+1}.
\end{equation}

For the orientation-based metrics, we first calculated the rotation difference between consecutive frames:
\begin{equation}
     \Delta \mathbf{Q}_{j,j+1}=\mathbf{Q}_{j+1}\mathbf{Q}_{j}^{-1},
\end{equation}
where $\mathbf{Q}_{j}$ and $\mathbf{Q}_{j+1}$ are unit quaternions representing the orientation of the frames. $\Delta \mathbf{Q}_{j,j+1}$ is a unit quaternion and thus can be referred to as rotation around the axis $\mathbf{\hat{k}}$ ($\mathbf{\hat{k}}=[k_x,k_y,k_z]$) by $\Delta \theta_{j,j+1}$ \cite{kellyMobileRoboticsMathematics2013} (Fig. \ref{fig_transformation}(b)):
\begin{eqnarray}
    \Delta \mathbf{Q}_{j,j+1}&=&[q_1,q_2,q_3,q_4]=\\ \nonumber
    & &[cos(\frac{\Delta \theta_{j,j+1}}{2}),\mathbf{\hat{k}}sin(\frac{\Delta \theta_{j,j+1}}{2})].
\end{eqnarray}
We calculated the angle $\Delta \theta_{j,j+1}$ (Fig. \ref{fig_transformation}(b)), which represents the orientation change between pairs of sampled frames:
 \begin{equation}
      \Delta \theta_{j,j+1}=2cos^{-1}(q_1),
 \end{equation}
where  $q_1$ is the first component of the quaternion  $\Delta \mathbf{Q}_{j,j+1}$.

For each participant and segment, outlier angle values were defined as angle values that were 35 times larger than the average of the angles across all the trials of that participant and segment. The source of the outliers is attributed to problems during the recording of the data. The entire segment that included an outlier angle was removed from the analysis. In the teleoperated condition of the Dry Lab dataset, this outlier removal procedure resulted in the removal of 6 segments. In the open condition of the Dry Lab dataset, and in the Porcine dataset, none of the segments were removed.

The \textit{angular displacement}, was defined as:
\begin{equation}
    A=\sum_{j=1}^{N-1} |\Delta \theta_{j,j+1}|.
\end{equation}
This metric quantifies the overall change in orientation during the movement -- the angular path. Note that this metric is different to the angular path metric used to assess laparoscopic skill, which refers to the amount of rotation around only one or two of the tool’s axes \cite{hogleDocumentingLearningCurve2007,hofstadStudyPsychomotorSkills2013}. In the open needle-driving, we measured two orientations -- one for each magnetic tracker. Because the trackers were rigidly attached to the driver, we assumed that as long as the needle-driver is closed around the needle, the change in the orientation between subsequent samples should be equal for both of the trackers. However, some participants held the driver so that one of their fingers touched one of the trackers. This contact disturbance caused movements of the tracker, and therefore, unintentional changes in the orientation that could inflate the angular displacement metric. Therefore, we calculated the angular displacement for the two trackers, and used the smaller angular displacement in further calculations.

The \textit{rate of orientation change} was defined as:
\begin{equation}
    C=\frac{1}{N-1}\sum_{j=1}^{N-1} \frac{|\Delta \theta_{j,j+1}|}{\Delta t_{j,j+1}},
\end{equation}
where $\Delta t_{j,j+1}$ is the time difference between the subsequent samples. This metric quantifies the rate of the change of the instrument orientation during the movement. In the open needle-driving, we calculated $\Delta \theta_{j,j+1}$ from the same tracker that was used for the calculation of the normalized angular displacement (without the finger contact disturbance).

\subsection{Statistical analysis}
\subsubsection{Dry Lab Dataset}\hfill\\
In this study we focused on differences between the performance of experienced surgeons and nonmedical users, and between the beginning and the end of the experiment. We did not perform statistical comparisons between the teleoperated and open conditions. Therefore, the following process was carried out for each condition separately.

For each trial, we calculated the four metrics for the first and second segments (I-transport and II-insertion). We log-transformed the metrics to correct their non-normal distributions. We calculated the averages of the first and last 10 trials of each participant for each metric and each segment. For each metric, we fit a 2-way ANOVA model with repeated measures on one factor (mixed model ANOVA), with expertise (experienced surgeon / nonmedical user) as the between-participants factor, and trial (early/late) as the within-participant factor. We used Bonferroni correction for post-hoc comparisons.

\subsubsection{Porcine Dataset}\hfill\\
 We calculated the values of the four metrics for each insertion attempt for each of the seven participants. We then calculated the average value of each metric per participant. Following that, we calculated the average value of each metric for each expertise group (experienced surgeons and novice surgeons), and the difference between the groups' averages ($\Delta_{Exp.-Nov.}$). The sample size of this dataset is small, and therefore, we used permutation tests \cite{goodPermutationTestsPractical2013} to test the statistical significance of the difference between the different expertise groups. We reassigned the seven participants' averages into the 35 possible combinations of groups: one group with three participants (`experienced surgeons'), and a second group with four participants (`novice surgeons'). For each combination we calculated ($\Delta_{Exp.-Nov.}$). To calculate the significance, we counted the number of combinations in which ($\Delta_{Exp.-Nov.}$) was equal to or higher in absolute value than the original ($\Delta_{Exp.-Nov.}$), and divided it by the number of the possible combinations.
 
Statistical significance was determined at the 0.05 threshold. We used the Matlab Statistics Toolbox for our statistical analysis.

\section{Results}
\subsection{Comparing needle-driving performance of experienced surgeons and nonmedical users in the Dry Lab dataset}

\begin{figure*}[!t]
\centering
\includegraphics[width=\textwidth]{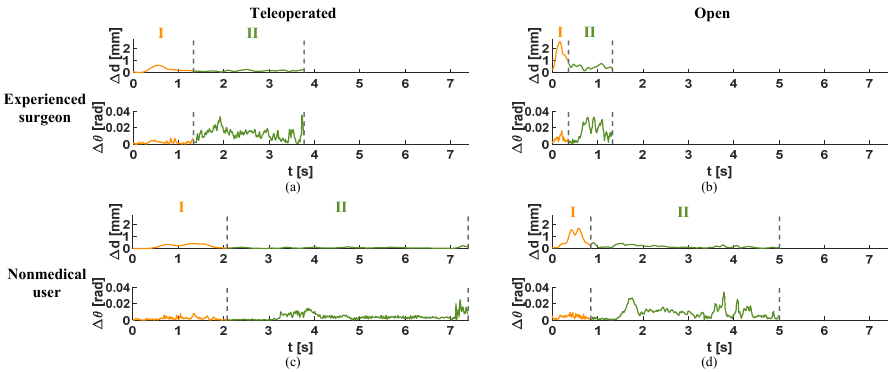}
\caption{Examples of $\Delta d$ and $\Delta \theta$ in teleoperated and open trials of an experienced surgeon and a nonmedical user from the Dry Lab dataset. (a) Trial of an experienced surgeon in the teleoperated session. (b) Trial of an experienced surgeon in the open session. (c) Trial of a nonmedical user in the teleoperated session. (d) Trial of a nonmedical user in the open session.}
\label{fig_example_trial}
\end{figure*}

Fig. \ref{fig_example_trial} depicts examples of a teleoperated and an open trial of an experienced surgeon in the upper panels, and a nonmedical user in the lower. Qualitatively, it is evident that the experienced surgeon completed both tasks faster than the nonmedical user, and with higher rate of orientation change. Figs. \ref{fig_results_tele}-\ref{fig_results_open} depict the metrics in the first two segments, for the teleoperated (Fig. \ref{fig_results_tele}) and the open (Fig. \ref{fig_results_open}) conditions, as a function of trial number (left panels), and the averages of the first and the last 10 trials (right panels). Most of the noticeable differences between experienced surgeons and nonmedical users are in segment II (insertion). This observation is not surprising because the driving of the needle through the tissue (segment II) is the challenging aspect of the task. Nevertheless, for completeness, we briefly present the full analysis of segment I and then focus on segment II.

\subsubsection{Segment I-transport}\hfill\\
The statistical analysis of segment I (transport) showed that for most of the metrics, the differences between experienced surgeons and nonmedical users, in both conditions (teleoperated and open) were not statistically significant. The only metric that showed statistically significant differences between experienced surgeons and nonmedical users was task time (teleoperated: $F_{1,13}=8.942, p=0.010, \Delta_{Exp.-Non.}=-0.476$, open: $F_{1,13}=6.206, p=0.027, \Delta_{Exp.-Non.}=-0.316$). In addition, the improvement between early and late trials was statistically significant for the task time in both conditions (teleoperated: $F_{1,13}=10.147, p=0.007, \Delta_{Late-Early}=-0.463$, open: $F_{1,13}=14.003, p=0.002, \Delta_{Late-Early}=-0.256$), and for the rate of orientation change in the open condition ($F_{1,13}=7.891, p=0.015, \Delta_{Late-Early}=0.223$). 

\subsubsection{Segment II-insertion}\hfill\\
Table I summarizes the results of the mixed effects ANOVA for the different metrics. Post-hoc comparisons are presented only when the interaction expertise*trial was significant. 

\begin{figure}[!t]
\centering
\includegraphics[width=3.5in]{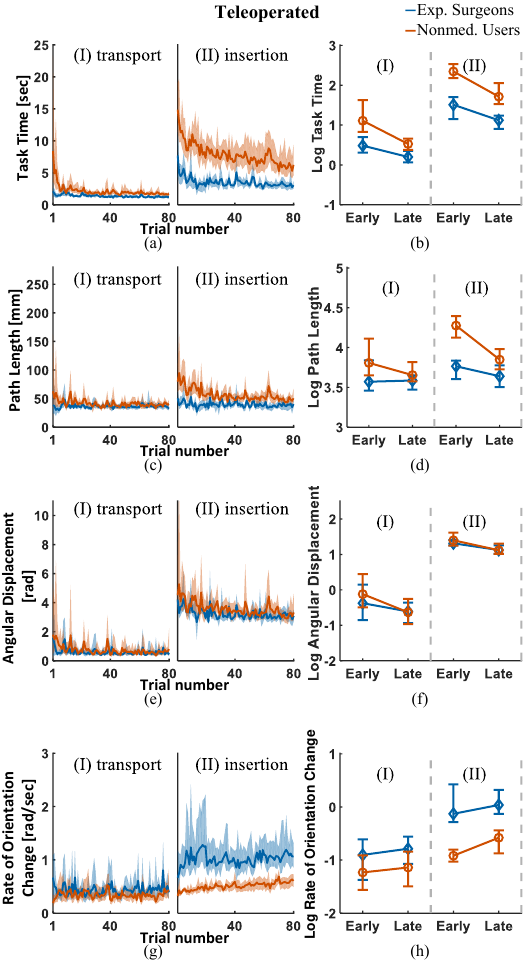}
\caption{The four metrics in the first two segments for the teleoperated condition of the Dry Lab dataset. Left panels -- the metrics as a function of trial number. Lines are means, shaded areas are 95\% bootstrap confidence intervals. Right panels -- average metric in the first 10 (early) and last 10 (late) trials. Markers are means, error bars are 95\% bootstrap confidence intervals.}
\label{fig_results_tele}
\end{figure}

\begin{figure}[!t]
\centering
\includegraphics[width=3.5in]{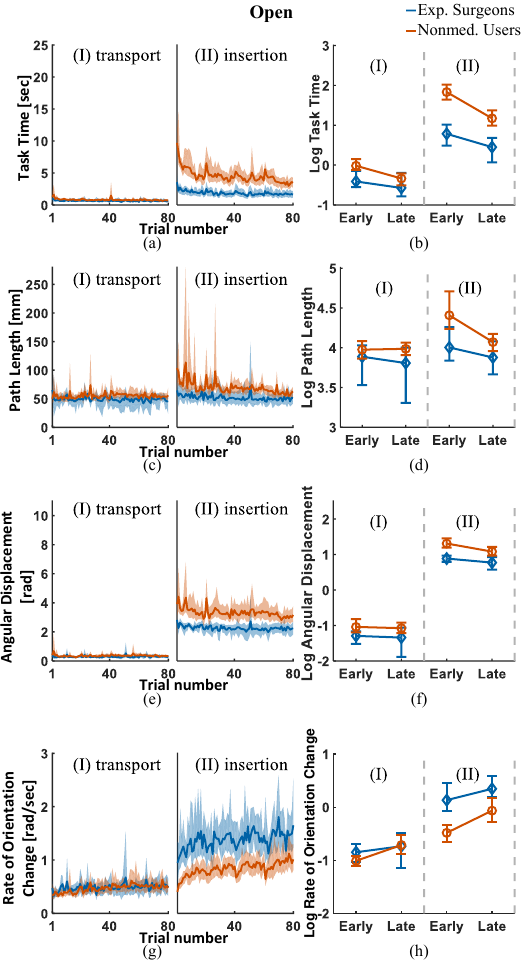}
\caption{The four metrics in the first two segments for the open condition of the Dry Lab dataset.  Left panels -- the metrics as a function of trial number. Lines are means, shaded areas are 95\% bootstrap confidence intervals. Right panels -- average metric in the first 10 (early) and last 10 (late) trials. Markers are means, error bars are 95\% bootstrap confidence intervals.}
\label{fig_results_open}
\end{figure}

\textit{Task time} is a classical metric, and we expected shorter task times for more experienced surgeons. Indeed, the task time of the experienced surgeons was shorter than of nonmedical users, and in the last trials of the experiment, task time was shorter than in the first trials (Fig. \ref{fig_results_tele}(a-b), \ref{fig_results_open}(a-b)). This observation is supported by the statistical analysis -- for both conditions (teleoperated and open), the effect of expertise and trial was statistically significant. In the open condition, the interaction between trial and expertise was statistically significant, because the improvement of the nonmedical users was greater than of experienced surgeons. Nevertheless, the difference between them remained statistically significant even in the last trials. 

 \textit{Path length} is related to the classical economy of motion metrics. Fig. \ref{fig_results_tele}(c-d) and \ref{fig_results_open}(c-d) show that in segment II, in both conditions (teleoperated and open), experienced surgeons had a shorter path length than nonmedical users, and that there was an improvement between early and late trials. These effects were statistically significant. In the teleoperated condition, there was a statistically significant interaction between expertise and trial. In the early trials, the paths of experienced surgeons were shorter than of nonmedical users (Fig. \ref{fig_results_tele}(d)). The nonmedical users improved more than the experienced surgeons, and as a result, in the late trials, there was no longer a statistically significant difference in path length between experienced surgeons and nonmedical users. These results are consistent with our previously reported analysis of the entire task \cite{niskyTeleoperatedOpenNeedle2015}. The fact that there was no difference between experienced surgeons and nonmedical users after only 80 trials suggests that, at least in some tasks, this metric alone is insufficient for surgical skill assessment. 
 
\textit{Angular displacement.} Our task requires rotation of the needle along its arc for successful insertion into the tissue. Therefore, we hypothesized that a large angular displacement will be correlated to surgical experience. However, Fig. \ref{fig_results_tele}(e-f) depict that in the teleoperated condition, there is no statistically significant difference between experienced surgeons and nonmedical users. Moreover, Fig. \ref{fig_results_open}(e-f) show that in the open condition, experienced surgeons had a statistically significant smaller angular displacement than nonmedical users. In addition, in the teleoperated condition, the angular displacement at the end of the experiment was statistically significantly smaller than in the early trials. 

A careful examination of Fig. \ref{fig_example_trial}(d) suggests a reason for these surprising results. The nonmedical user tried a few times unsuccessfully to rotate the needle through the tissue, and accumulated a large angular displacement that does not necessarily reflect a successful drive of the needle (panel b). This motivated us to propose a metric that quantifies the rate of orientation change rather than its accumulation.

\textit{Rate of orientation change.} Examination of orientation change trajectories ($\Delta \theta$) (Fig. \ref{fig_example_trial}) suggests that experienced surgeons perform the insertion in one attempt, and use faster orientation changes. Therefore, we hypothesized that a higher rate of orientation change will be correlated to surgical experience. Fig. \ref{fig_results_tele}(g-h) and \ref{fig_results_open}(g-h) show that in segment II, in both sessions (teleoperated and open), experienced surgeons changed their orientation faster than nonmedical users, and that in the last trials of the experiment, the rate of orientation change was higher than in the first trials. The statistical analysis supported this observation, and showed that for both conditions (teleoperated and open), the effect of expertise and trial was statistically significant.

\begin{table}[!t]
\renewcommand{\arraystretch}{1.3}
\caption{Statistical Analysis Summary (Dry Lab -- segment II)}
\label{table_example}
\centering
\includegraphics[width=3.5in]{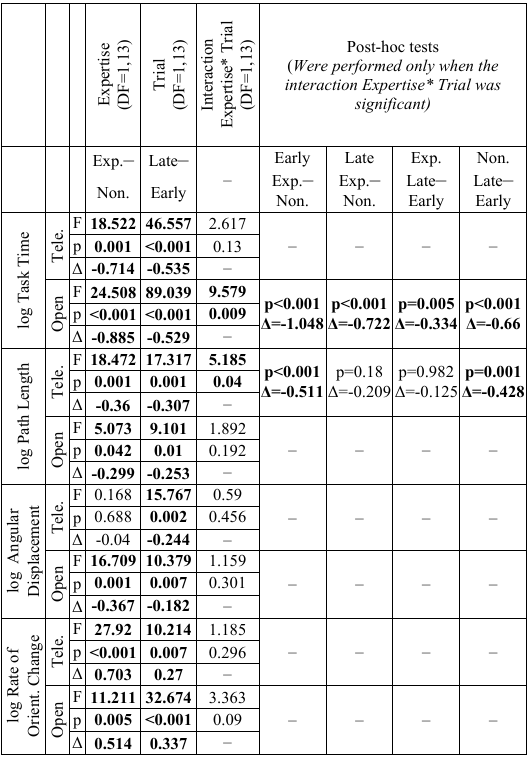}
\raggedright{Bold font indicates statistically significant effects.\\
Exp., Non., and Tele. are abbreviations for experienced surgeons, nonmedical users, and teleoperated, respectively.}
\end{table}

\subsection{Comparing performance of experienced and novice surgeons in the Porcine dataset}
\begin{figure}[!t]
\centering
\includegraphics[width=3.5in]{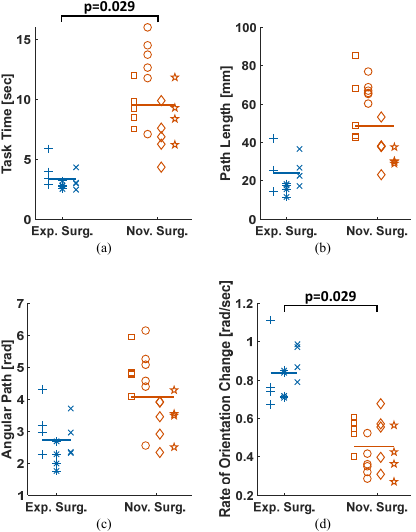}
\caption{The four metrics in the insertion segment of the Porcine dataset. The different markers indicate the seven participants, and the horizontal lines represent the average value of the metric across the participants. Exp. Surg. and Nov. Surg. are abbreviations for experienced surgeons and novice surgeons, respectively. The p values are presented for the metrics in which the permutation test was statistically significant.}
\label{ClinicalLikeResults}
\end{figure}

Fig. \ref{ClinicalLikeResults} depicts the results of the metrics in the insertion segments of the suturing task. The experimental conditions in this dataset are different to those in the Dry Lab dataset, i.e., novice surgeons instead of nonmedical users, and the training is on a porcine model instead of a dry lab task. Despite these differences,  the results of the two datasets are consistent with each other. There are large and statistically significant differences between experienced surgeons and novice surgeons in  \textit{task time} ($\Delta_{Exp.-Nov.}=-6.1442 [sec], p=0.029$) and \textit{rate of orientation change} ($\Delta_{Exp.-Nov.}=0.38157 [rad/sec], p=0.029$). The differences in \textit{path length} ($\Delta_{Exp.-Nov.}=-24.4987 [mm], p=0.086$) and \textit{angular displacement} ($\Delta_{Exp.-Nov.}=-1.3479 [rad], p=0.086$) are less pronounced. While these results are very promising, and demonstrate the performance our new rate of orientation change metric on more realistic data, we are cautious when it comes to drawing clear conclusions, due to the small size of this dataset.

\section{Discussion}
In this study, we developed two orientation-based metrics for surgical skill evaluation. These metrics capture critical aspects of angular motion that are not taken into account in other motion-based metrics which are calculated using the position of the tools. We tested the new metrics using two datasets, which differ in their conditions. The Dry Lab dataset includes data of a very structured and simplified task, which was performed by experienced surgeons and nonmedical users. Using this dataset, we were able to test our new metrics under controlled conditions. The Porcine dataset includes data of training on a porcine model, which was performed by experienced and novice surgeons. Using this dataset, we demonstrated the ability of our metrics on more realistic situations. We tested our new metrics alongside task time and path length metrics, which are classical metrics for surgical skill evaluation. 
This allowed us to examine the performance of our metrics alongside the performance of metrics that are commonly used today. 
Using our new metric (the rate of orientation change) we found that experienced surgeons change the tool’s orientation statistically significantly faster than nonmedical users / novice surgeons. This result is consistent with the statistically significant difference we found using the classical task time metric. Surprisingly, using our second new metric (angular displacement) we found that experienced surgeons have shorter angular paths than nonmedical users / novice surgeons. This difference was statistically significant only for the open technique. Taken together, our results suggest that when assessing skill in procedures that require control of orientation, in addition to the existing metrics, it is important to use orientation-based metrics. 

\subsection{Task segmentation}
The needle-driving and suturing tasks include several segments. Each segment in these tasks has different requirements in terms of task constraints, and may require the use of different metrics to assess surgical skills. For example, the needle transport segment probably does not require prominent orientation change. Therefore, prior to metrics calculation, we used characteristics of the movement (Dry Lab) or video data (Porcine) to segment it. Our results indicate that the segmentation was important -- for example, in the Dry Lab dataset, the angular displacement was much higher during insertion (segment II) than during transport (segment I). Additionally, the rate of orientation change revealed differences between experienced surgeons and nonmedical users during insertion, but not during the transport segment. These results highlight the importance of segmentation in surgical skill assessment. 

In the Dry Lab dataset, the segments were part of the design of the experiment, and therefore, their definition was simple. In most of the clinical procedures, segmentation is also very important \cite{n.ahmidiDatasetBenchmarksSegmentation2017}, and exists both on a macro and on a micro level. For example, in prostatectomy or thymectomy the procedure can be segmented into discrete steps: anatomical structures dissection, removal of anatomy of interest, and anastomosis. Each of these steps can be further segmented into sub-movements, like the segments detailed within the Dry Lab dataset. To address the segmentation challenge, several prior studies developed algorithms for surgical task segmentation \cite{jamesEyegazeDrivenSurgical2007,n.ahmidiDatasetBenchmarksSegmentation2017,menegozzoSurgicalGestureRecognition2019, itzkovichUsingAugmentationImprove2019}. 

In the Dry Lab dataset, we focused the majority of our analysis on the insertion of the needle, because this was the most challenging aspect of the task, and because most of the differences between experienced surgeons and nonmedical users were observed in that segment. Additionally, in the Porcine dataset, to be able to compare between the movements we chose to focus on the insertion part, which was consistent across participants and attempts. Therefore, the remainder of the discussion focuses on the insertion segment.

\subsection{Metrics}
Consistently with previous studies \cite{vemerMeasurementsLevelSurgical2003,hofstadStudyPsychomotorSkills2013,vansickleConstructValidationProMIS2005,chandraComparisonLaparoscopicRobotic2010,tauschContentConstructValidation2012}, we found that experienced surgeons completed the task faster than nonmedical users. However, the speed-accuracy tradeoff -- the inverse relation between the speed of the movement and its accuracy \cite{fittsInformationCapacityHuman1954,chienAccuracySpeedTrade2010} suggests that surgeons may compromise accuracy to complete the task very fast. Therefore, task time must be accompanied by accuracy metrics \cite{moorthyObjectiveAssessmentTechnical2003,bannMeasurementSurgicalDexterity2003}.

Although path length is a common measure for surgical skill, there is disagreement regarding its effectiveness. Several studies showed that path length is a useful metric \cite{hofstadStudyPsychomotorSkills2013,vansickleConstructValidationProMIS2005,kenneyFaceContentConstruct2009}, but others found it to be less adequate \cite{dattaUseElectromagneticMotion2001,chmarraRetractingSeekingMovements2008}. For example, during blunt tissue dissection, it is common for novices to be too `timid' and do inefficient and small instrument sweeps to separate tissue planes, whereas experienced surgeons, who understand tissue tolerances better, may make much broader sweeping motions, thus elevating overall path length. Our results show difference between the path length of experienced surgeons and novices in the Dry Lab dataset, but at the end of the teleoperated sessions, this difference was not statistically significant. Similarly, we observed differences in path length between the expertise groups in the Porcine dataset, but not in all the surgeons. The difference between the averages of the two expertise groups was not statistically significant. Therefore, we believe that, at least in needle-driving, path length alone is insufficient for quantifying surgical skill.

To quantify the range of the tool's orientation change, we proposed a new metric of angular displacement. We expected that experienced surgeons will have a larger angular displacement. Surprisingly, in the teleoperated condition of the Dry Lab dataset and in the Porcine dataset, we found no significant difference, and in the open condition, we found differences in the opposite direction to what we expected. A possible explanation is corrections of the tool's orientation to enable a better insertion. Nonmedical users probably used many such (unsuccessful) correction attempts that resulted in a large total angular displacement. On the other hand, the experienced surgeons knew exactly how to rotate their hand as required, and needed fewer corrections. Therefore, a movement of an experienced surgeon with fast large accumulated orientation, and a movement of a nonmedical user with many corrections may yield the same angular displacement. Our results are in agreement with a previous study of suturing skill in a virtual reality simulator \cite{kazemiAssessingSuturingTechniques2010}. They found that during needle insertion, trained participants had less orientation change than untrained participants, and suggested that this result may be due to errors in needle grasping and penetration angle.

The last metric (rate of orientation change) is not affected by the accumulation of unsuccessful attempts of insertion. Indeed, it showed statistically significant differences between nonmedical users and experienced surgeons in the Dry Lab dataset. We also found that in the last trials, the participants changed the tool's orientation faster than in the first trials. Similarly, in the Porcine dataset this metric, together with task time, was the most successful in characterising the differences between expertise levels. These results demonstrate that the rate of the change of the tool's orientation is important for the success of needle-driving. 

Our ultimate goal is to develop new metrics for differentiation between expertise levels. Our results of statistically significant differences between the rate of orientation change of experienced surgeons and nonmedical users / novice surgeons fall short of
proving this differentiation. However, the  size of the effect is substantial -- the experienced surgeons are almost twice as fast in orientation change compared to nonmedical users / novice surgeons in both datasets. Moreover, the differences are statistically signficant even in a small sample size. This suggests that our new metric presents a promising step towards the eventual differentiation between expertise levels.

For each surgical task it is important to choose the relevant metrics. For example, each exercise of the da Vinci Skills Simulator (Intuitive Surgical, Inc.) has different requirements and therefore, each exercise has a unique scoring method \cite{VinciXiSkills2018}. The new orientation-based metric may help get a more accurate estimation of technical skill in tasks that involve control of orientation, such as suturing. Each individual metric has its strengths and its limitations. Moreover, it appears necessary to combine more than one metric. For example, in a task of needle insertion, if only orientation-based metrics are used, it is possible to `game' the task by significantly and quickly rotating the tool before starting the insertion and getting a better score. Therefore, in developing training curricula it is important to combine many metrics for skill assessment.

Because the rate of orientation change quantifies characteristics that can be explained, trainers will be able to give trainees informative feedback on how to improve their movements. For example, they will be able to guide trainees to rotate the tool faster during specific parts of the task. In addition, it is possible to tailor haptic assistance or resistance training for the control of orientation \cite{m.m.coadTrainingDivergentConvergent2017,oquendoRobotAssistedSurgicalTraining2019}. For example, future studies can test whether haptic guidance that will rotate the hand of the participant faster during surgical tasks or, alternatively, adding a resistance to rotation can increase the rate of orientation change.

\subsection{Implications to human motor control}
The insertion of the needle involves a complicated motion that requires control of the tool's orientation. In human motor control, point-to-point and planar drawing movements have been well studied, and many models were proposed to explain how we control these movements \cite{lacquanitiLawRelatingKinematic1983,flashCoordinationArmMovements1985,leibMinimumAccelerationConstraints2012}. Three-dimensional movements were studied to a much lesser extent \cite{maozComplexUnconstrainedThreedimensional2009,sharonExpertiseTeleoperationTask2018}, and the control of orientation \cite{fanControlHandOrientation2006} is almost never explored. 

In addition, our task involves insertion of a curved needle into either artificial or real tissue. Interaction with elastic objects is often studied in one-dimensional movements \cite{leibEffectForceFeedback2015} and needle insertion into soft tissue was also previously studied \cite{niskyPerceptionActionTeleoperated2011}, but only using a simplified model of tissue in a constrained task. However, models of movement and orientation coordination in three dimensional movement while manipulating complex end-effectors (such as our needle) are yet to be developed. Our new orientation-based metrics may help in understanding how surgeons control the orientation of their hands and instruments. Therefore, in future work, our study can advance the understanding of movement coordination in realistic scenarios. 

\subsection{Limitations and Future work}
A video of the experiments in the Dry Lab dataset was not recorded, and therefore, we were not able to segment the movement after the two first segments. An analysis of the last two segments could add valuable information, and could contribute to the examination of the new metrics that we proposed. However, the first two segments require insertion of the needle through the tissue, which is a complex movement for inexperienced surgeons. Therefore, we believe that we can learn about surgical skill, and examine our new metrics using the first two segments.

The needle-driving task of the Dry Lab dataset does not represent a real situation. For example, while surgeons often use the thread to position the needle, in the Dry Lab task there was no thread. To teach and evaluate basic surgical skills, it is typical to simplify real situations \cite{caccianigaEvaluationInanimateVirtual2020,gaoJHUISIGestureSkill2014}, or even to use tasks which are not related to real surgical situations (e.g., peg transfer \cite{smithFundamentalsRoboticSurgery2014}). To test the potential of our new metrics in realistic situations, we examined them on the Porcine dataset. Although this dataset is small, the results suggest that these metrics may also be used in more realistic tasks. 

For the rate of orientation change, there was great variability within the experienced surgeons group of the Dry Lab dataset (Fig. \ref{fig_results_tele}(g) and Fig. \ref{fig_results_open}(g)). This may be a result of different strategies, or different skill levels within the group. A composite of all the metrics may provide more granular discrimination among surgeons -- not just novices and experienced surgeons -- but novice to intermediate and intermediate to expert and all levels between. Future studies with additional participants from different expertise groups such as medical students, residents, fellows, and experienced surgeons with a larger variety of case experience are needed to explore such composite metrics. 

Correlating the new orientation-based metric with global rating scores such as OSATS may add validation to these novel performance metrics. In the main dataset that we used in this study, the Dry Lab, we do not have  video recordings of the experiment, and therefore, we cannot extract global rating scores of the movements. However, we compare between experienced surgeons and engineering students, which are two very different groups. This maximizes the possible differences in expertise, and facilitates development of new metrics. To validate this metric, further investigation is needed, including comparisons with global rating scores. In the Porcine dataset, we also do not have these ratings, and there are too few surgeons in the Porcine dataset for meaningful conclusions using such ratings.

In the Dry Lab dataset, we compared between experienced surgeons and engineering students; medical residents were not included. The needle-driving task does not require clinical judgment, and therefore is less reliant on clinical knowledge and training. Moreover, medical residents early in their residency have very limited suturing experience, and hence they resemble engineering students in terms of suturing skills. Therefore, comparing between these two groups would allow us to test the performance of the new metrics we developed. Additionally, testing our metrics on the Porcine dataset, which consisted of surgeons with different expertise levels, yielded similar results.

\section{CONCLUSIONS}
We developed two new metrics for surgical skill evaluation. The rate of orientation change showed promising results. This metric captures technical aspects of the rotation of the hands and instruments that are taught during surgical training and had not been quantified by any other metric. We demonstrated that the rate of orientation change correlates with experience in both teleoperated and open needle insertion on a dry lab model, as well as on a porcine model. In addition, our results highlighted the importance of evaluating each segment of the movement separately. Future studies are needed to test this metric on a larger cohort of surgeons, and to translate kinematic metrics into meaningful training feedback to facilitate more efficient training. Characterizing the movements of the surgeons may help improve the evaluation and the acquisition of motor skills that are critical to surgery, and may also provide new insight into how to improve the control of surgical robots, and the training of new surgeons.

\section*{Acknowledgment}
We thank Allison Okamura, Michael Hsieh, Zhan Fan Quek and Yuhang Che for providing the experimental data of the Dry Lab dataset.

\ifCLASSOPTIONcaptionsoff
  \newpage
\fi



\bibliographystyle{IEEEtran}
\bibliography{Refs}
%
 
%





\end{document}